%% file: lrec2026.tex
\documentclass[10pt, a4paper]{article}

\usepackage[final]{lrec2026} 

\usepackage{times}
\usepackage{latexsym}
\usepackage[T1]{fontenc}
\usepackage[utf8]{inputenc}
\usepackage{microtype}
\usepackage{inconsolata}
\usepackage{graphicx}
\usepackage{enumitem}
\usepackage{xcolor}
\usepackage{soul}
\usepackage{multirow}
\usepackage{array}
\usepackage{lscape}
\usepackage{svg}
\usepackage{tcolorbox}
\usepackage{amsmath}
\usepackage{subcaption}
\usepackage{lineno}
\usepackage{longtable}
\usepackage{caption}
\usepackage{subcaption}
\usepackage{xurl}
\usepackage{float}
\usepackage{stfloats}

\usepackage[utf8]{inputenc}
\usepackage[linesnumbered,ruled,vlined]{algorithm2e}
\usepackage{amsmath}

\usepackage{booktabs}        
\usepackage{tabularx}        
\usepackage{adjustbox}       

\usepackage{siunitx}

\newcommand{\PreserveBackslash}[1]{\let\temp=\\#1\let\\=\temp}
\newcolumntype{R}[1]{>{\raggedleft\arraybackslash}p{#1}} 
\newcolumntype{L}[1]{>{\raggedright\arraybackslash}p{#1}} 
\newcolumntype{C}[1]{>{\centering\arraybackslash}p{#1}} 

\usepackage{caption}
\usepackage{colortbl}

\title{LLM-as-an-Annotator: Training Lightweight Models with LLM-Annotated Examples for Aspect Sentiment Tuple Prediction}

\name{Nils Constantin Hellwig$^1$, Jakob Fehle$^1$, Udo Kruschwitz$^2$, Christian Wolff$^1$}

\address{
$^1$Media Informatics Group, University of Regensburg, Regensburg, Germany \\
$^2$Information Science Group, University of Regensburg, Regensburg, Germany \\
nils-constantin.hellwig@ur.de, jakob.fehle@ur.de, udo.kruschwitz@ur.de, christian.wolff@ur.de
}

\input{00_abstract}

\begin{document}

\maketitleabstract

\input{01_introduction}
\input{02_related_work}
\input{03_methodology}
\input{04_results}
\input{05_discussion}
\input{06_conclusion}
\input{07_ethical_considerations}
\nocite{*}
\section{Bibliographical References}\label{sec:reference}

\bibliographystyle{lrec2026-natbib}
\bibliography{lrec2026-example}

\input{08_appendix}

\end{document}

%% file: 00_abstract.tex
\abstract{
Training models for Aspect-Based Sentiment Analysis (ABSA) tasks requires manually annotated data, which is expensive and time-consuming to obtain. This paper introduces LA-ABSA, a novel approach that leverages Large Language Model (LLM)-generated annotations to fine-tune lightweight models for complex ABSA tasks. We evaluate our approach on five datasets for Target Aspect Sentiment Detection (TASD) and Aspect Sentiment Quad Prediction (ASQP). Our approach outperformed previously reported augmentation strategies and achieved competitive performance with LLM-prompting in low-resource scenarios, while providing substantial energy efficiency benefits. For example, using 50 annotated examples for in-context learning (ICL) to guide the annotation of unlabeled data, LA-ABSA achieved an F1 score of 49.85 for ASQP on the SemEval Rest16 dataset, closely matching the performance of ICL prompting with Gemma-3-27B (51.10), while requiring significantly lower computational resources.
 \\ \newline \Keywords{Aspect-Based Sentiment Analysis, Large Language Models, Data Annotation} }

%% file: 01_introduction.tex
\section{Introduction}

Sentiment Analysis (SA) involves recognizing the emotional mood expressed in a text, for example in domains like e-commerce or social media \citep[p.~7]{liu2022sentiment}. Traditional sentiment classification assigns a single polarity to a text, capturing its overall emotional tone \citep[p.~59]{liu2022sentiment}. In contrast, recent research considered more granular tasks, such as Target Aspect Sentiment Detection (TASD) and Aspect Sentiment Quad Prediction (ASQP), which fall under Aspect-based Sentiment Analysis (ABSA) \citep{zhang2021aspect, negi2024hybrid, lietal2024majority, liu2024chatasu}. These tuple prediction tasks enable a detailed extraction of opinion structures, enhancing the analysis of nuanced sentiments in diverse linguistic contexts.  TASD involves identifying (aspect term \textit{a}, aspect category \textit{c}, sentiment polarity \textit{p})-triplets, whereas ASQP allows for an even more precise insight by additionally predicting the opinion term \textit{o}, resulting in quadruples \citep{zhang2021aspect}. For example, given the sentence \textit{``The pizza was tasty.''}, the corresponding ASQP annotation would be (\textit{a}: \textit{pizza}, \textit{c}: \textit{food general}, \textit{o}: \textit{tasty}, \textit{p}:  \textit{positive}).

However, the increased complexity of these tasks comes at the cost of data annotation: creating labelled datasets for TASD and ASQP is even more labour-intensive than for traditional sentiment tasks, requiring annotators to identify multiple interrelated text spans, the related aspect category and sentiment polarities \citep{nasution2024chatgpt,negi2024hybrid,wang2024generative}. As a result, annotated resources for these tasks are often limited in scale and domain coverage.

Large Language Models (LLMs) such as Gemma-3 \citep{team2025gemma}, LLaMA-3.1 \citep{grattafiori2024llama} and GPT-4 \citep{achiam2023gpt} recently demonstrated strong performance in both zero-shot and few-shot settings, showing the ability to generalize well from task instructions and limited examples \citep{bai2024compound,gou2023mvp,hellwig2025we}. 

Despite their considerable performance, LLMs are resource-intensive, requiring substantial computational power and memory \citep{xu2025ds2absa}. This poses challenges for practical deployment, especially at scale or in scenarios involving privacy-sensitive data, particularly when relying on external APIs offering access to an LLM \citep{chen2025empirical, li2025easyjudge}.

While zero-shot and few-shot methods are effective in scenarios without labelled data, they do not exploit the potentially available unlabelled textual resources. An alternative approach, which has been employed across various NLP tasks, involves using LLMs to annotate unlabelled data and subsequently leveraging these annotations to train smaller, more computationally efficient models \citep{huang2022large,malik2024pseudo,yang2024self, zhang2023llmaaa}.

In this study, we evaluate whether LLMs can be leveraged as data annotators for aspect sentiment tuple prediction, specifically, ASQP and TASD. We introduce \underline{L}LM-as-an-\underline{A}nnotator (LA-ABSA), a novel approach that labels unlabelled examples using Gemma-3-27B and subsequently trains lightweight language models\footnote{Language models with fewer than 1 billion parameters were considered as lightweight.} on the LLM-annotated examples. We systematically evaluate this approach by addressing three key research questions:

\begin{itemize}
    \item \textbf{RQ1:} Do lightweight language models trained on LLM-annotated examples perform as well as those trained on expert-annotated data?
    \item \textbf{RQ2:} Do lightweight language models trained on LLM-annotated examples perform better than zero-shot and few-shot prompted LLMs?
    \item \textbf{RQ3:} Does LA-ABSA outperform data augmentation methods?
\end{itemize}

Our main contributions are as follows:

\begin{itemize}
    \item We introduce LA-ABSA, enabling lightweight language model training for aspect sentiment tuple prediction using LLM-generated annotations, while reducing dependency on expensive expert labelling.
    \item Evaluation across five datasets shows LA-ABSA achieves comparable performance to LLM prompting with lower energy consumption for large-scale deployments.
    \item We provide publicly available code (\url{https://github.com/NilsHellwig/LA-ABSA}) adaptable to other domains and languages.
\end{itemize}

%% file: 02_related_work.tex
\section{Related Work}

This section presents related work on two main aspects: (1) state-of-the-art (SOTA) methodologies for Target Aspect Sentiment Detection (TASD) and Aspect Sentiment Quad Prediction (ASQP), and (2) approaches aimed at mitigating annotation effort in Aspect-based Sentiment Analysis (ABSA).

Notably, all subsequently mentioned approaches for tackling tuple prediction tasks are primarily evaluated on the SemEval 2015 and 2016 datasets, which include TASD annotations \citep{pontiki2015semeval,pontiki2016semeval}. For the ASQP task, the extended versions of these datasets introduced by \citet{zhang2021aspect} were employed, which additionally include opinion term annotations.

\subsection{Fine-tuning Language Models for Tuple Prediction}

\textbf{Fine-Tuning Approaches.} For both ASQP and TASD, generative approaches achieved SOTA performance when small language models fine-tuned on annotated examples are employed. Most notably, Google's lightweight \texttt{T5-base} (223 million parameters) \citep{raffel2020exploring} text-to-text model. The approaches primarily differ in how tuples are represented. When using the \texttt{T5-base} model, Multi-view Prompting (MvP), introduced by \citet{gou2023mvp}, currently sets the SOTA performance for both ASQP and TASD on the Rest15 and Rest16 datasets, achieving, an F1 score of 72.76 on TASD-Rest16 and 60.39 on ASQP-Rest16. 

Close contenders include Dataset-level Order (DLO) \citep{hu2022improving} (TASD-Rest16: 71.79; ASQP-Rest16: 59.79) and Paraphrase \citep{zhang2021aspect} (TASD-Rest16: 71.97; ASQP-Rest16: 57.93). MvP is notably more computationally intensive than other methods, as it introduces element order-based prompts to guide tuple generation from multiple perspectives, meaning that different positional configurations of sentiment elements within a tuple are considered. This results in a multiplication of training samples by the number of permutations. Five permutations are considered for ASQP and TASD \citep{gou2023mvp}. In contrast, DLO selects only the top-3 permutations, while Paraphrase relies on a single fixed ordering \citep{zhang2021aspect}.

\textbf{Low-Rank Adaptation (LoRA) Fine-Tuning.} Fine-tuning LLMs with billions of parameters with Low-Rank Adaptation (LoRA)~\citep{hu2022lora}, a technique that updates only a small subset of parameters to reduce computational costs, has led to further improvements in F1 scores on both TASD and ASQP tasks. For example, \citet{vsmid2024llama} report F1 scores of 78.82 and 76.10 on the TASD-Rest16 dataset using Microsoft’s Orca-2-8B ~\citep{mitra2023orca} in its 13B and 7B variants, respectively. 

\subsection{Approaches for Minimizing Annotation Effort}

To address the challenge of limited labelled resources, three approaches have received particular attention in recent NLP research: (1) LLM prompting \citep{brown2020language, zhang2024sentiment, wang2025gpt, agarwal2024deep}, (2) traditional data augmentation methods such as EDA and back-translation \citep{imran2022data, sabty2021data, wei2019eda}, and (3) generative, LLM-based approaches that either augment existing annotated examples or generate new ones \citep{ding2024data, chung2023increasing, omura2024empirical, schmidt2024prompting}. This section examines their application and adaptation within ABSA.

\textbf{Zero-Shot and Few-Shot Learning.} To reduce annotation costs, recent studies explored zero-shot and few-shot learning for tuple prediction tasks \citep{bai2024compound,simmering2023large,vsmid2024llama,zhang2023sentiment,zhou2024comprehensive}. \citet{hellwig2025we} achieved SOTA in-context learning (ICL) performance on the ASQP and TASD tasks, reaching an F1 score of 66.03 on TASD-Rest16, though below SOTA fine-tuned performance slightly exceeding 70. However, their few-shot configurations outperformed established fine-tuning on TASD and ASQP tasks when only a few examples are given for training \citep{hellwig2025we,varia2023instruction}.

\textbf{Traditional Data Augmentation Methods.} Data augmentation for ABSA has been primarily evaluated in fine-tuning settings with full annotated datasets \citep{li2023data,liesting2021data,wang2023generative}. \citet{liesting2021data} applied Easy Data Augmentation (EDA) to aspect sentiment classification (aspect term + sentiment polarity), achieving modest gains of 0.5-1.0 points on SemEval datasets. EDA uses lexical transformations: synonym replacement, random insertion, swapping, and deletion \citep{wei2019eda}. Back-translation showed negligible improvements \citep{liesting2021data}.

\textbf{Generative Augmentation Strategies.} Leveraging the generative capabilities of language models, \citet{wang2024generative} propose a data augmentation approach for ASQP that augments new quads by exchanging opinion terms and sentiment polarity between tuples sharing the same aspect category, followed by training a quads-to-text (Q2T) model to generate diverse augmented texts based on the augmented quads. This approach yielded performance improvements of approximately 2 percentage points over the paraphrase-based method on both Rest15-ASQP and Rest16-ASQP datasets.\footnote{This approach was not adapted for our study, as we were unable to run the code provided on GitHub in its current form. Moreover, as reported in a GitHub issue, other users faced the same problem, and despite multiple attempts, we were unable to establish contact with the authors.}


Similarly, \citet{lu2025qaie} introduced an LLM-based approach named Quantity Augmentation and Information Enhancement (QAIE). QAIE uses PaLM (540 billion parameters) \citep{chowdhery2023palm} and assumes a small annotated set of examples with balanced distribution of \textit{k} examples per category. This approach demonstrated substantial performance improvements, achieving an F1 score increase from 22.97 to 35.31 (\textit{k=20}, 86 examples) on the ASQP-Rest15 dataset. QAIE enhances a given dataset by generating multiple paraphrased review texts with varied linguistic expressions and enriching them with additional implicit information before training a T5 model on the target task. The augmentation process generates more examples when both aspect and opinion terms are present in the source text, as this configuration enables the combination of synonym replacements and antonym substitutions for both opinion and aspect terms.

Finally, \citet{xu2025ds2absa} introduced DS\textsuperscript{2}-ABSA, a two-pronged data augmentation approach for E2E-ABSA (aspect term + sentiment polarity) that: (1) augments existing examples through sentence masking followed by paraphrasing of the masked instances, and (2) generates new training examples by first producing aspect and opinion terms, then synthesizing 20,000 complete examples based on these terms. Their approach outperformed other generative approaches adapted for E2E-ABSA.

%% file: 03_methodology.tex
\section{Methodology}

\begin{figure*}[t]
  \centering
  \includegraphics[width=\linewidth]{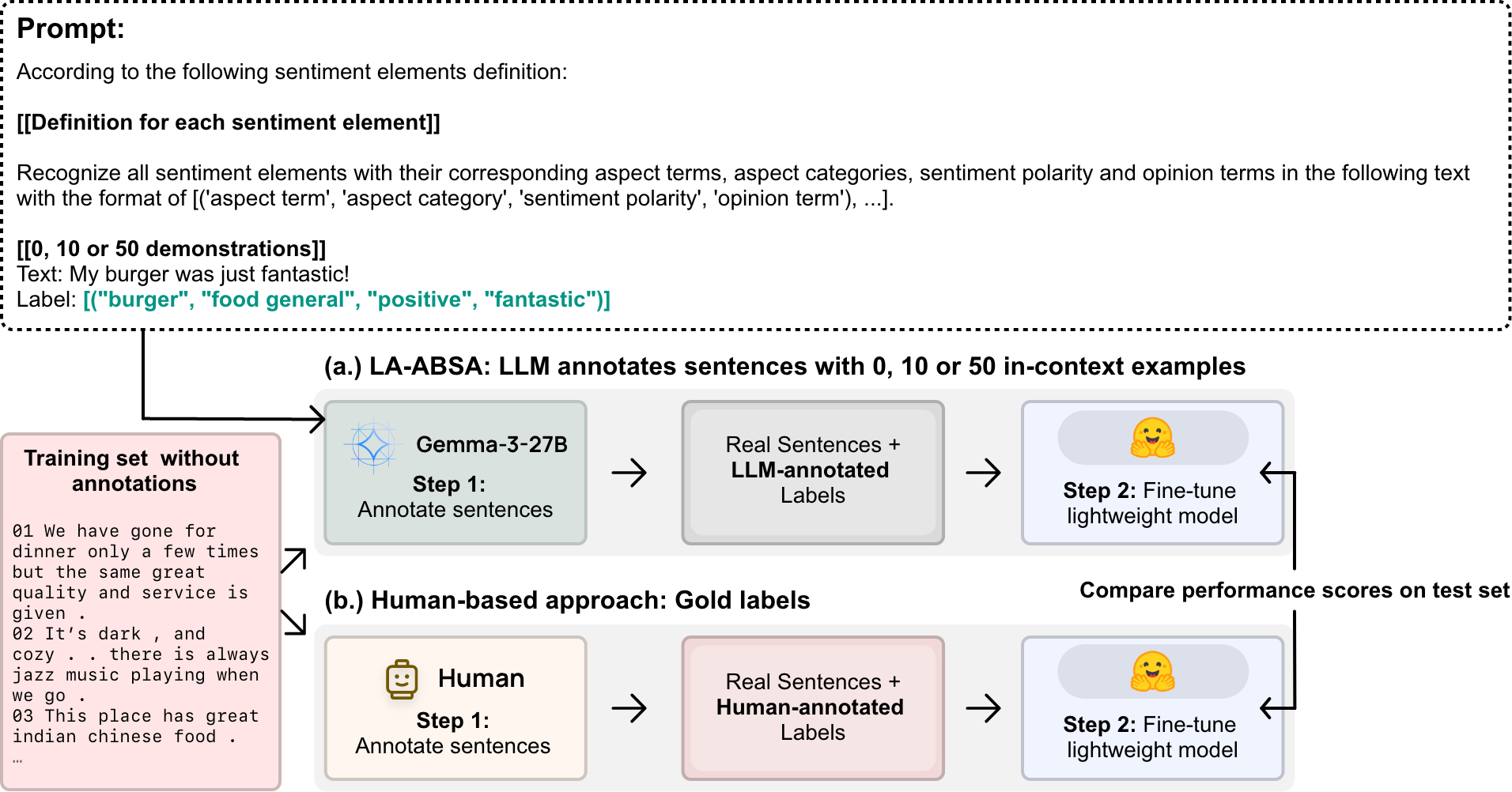}
  \caption{\textbf{Illustration of LLM-as-an-Annotator (LA-ABSA)}. A LLM (Gemma-3-27B) is prompted to annotate training examples, which are subsequently used to fine-tune lightweight state-of-the-art models for Target Aspect Sentiment Detection (TASD) and Aspect Sentiment Quad Prediction (ASQP).}
  \label{fig:graphical-abstract}
\end{figure*}

\subsection{LLM-as-an-Annotator (LA-ABSA)}

We introduce LA-ABSA, a novel approach for aspect sentiment tuple prediction (e.g., TASD or ASQP) that is dedicated to scenarios with limited annotated data. As illustrated in Figure \ref{fig:graphical-abstract}, our method leverages LLMs to automatically annotate unlabeled examples, thereby reducing the reliance on costly human annotation efforts.

\subsubsection{Annotator Module}\label{sec:annotator-module}

\begin{algorithm}[h]
\small
\DontPrintSemicolon
\SetAlgoLined
\KwIn{Unlabeled input set \(\mathcal{D} = \{x_i\}_{i=1}^N\), language model \(\mathcal{M}\), few-shot examples \(\mathcal{E} = \{(x_j, y_j)\}_{j=1}^k\)}
\KwOut{Annotated dataset \(\mathcal{D}_{\text{annotated}} = \{(x_i, y_i)\}_{i=1}^N\)}

\ForEach{\(x_i \in \mathcal{D}\)}{
    \(p_i := \texttt{construct\_prompt}(\mathcal{E}, x_i)\)\;
    \For{\(r = 1\) \KwTo 10}{
        \(y_i \leftarrow \mathcal{M}(p_i)\)\;
        \If{\texttt{validate\_label}(\(y_i\))}{
            \textbf{break}
        }
    }
    \If{\texttt{not} \texttt{validate\_label}(\(y_i\))}{
        \(y_i \leftarrow [\space]\)
    }
    \(\mathcal{D}_{\text{annotated}} \leftarrow \mathcal{D}_{\text{annotated}} \cup \{(x_i, y_i)\}\)
}
\Return \(\mathcal{D}_{\text{annotated}}\)
\caption{Annotator Module}
\label{alg:llm-fewshot-validation}
\end{algorithm}

To annotate the unlabeled training examples, we used the prompt template introduced by \citet{gou2023mvp}. The prompt comprises an explanation on all sentiment elements and the output format, few-shot examples (except for zero-shot conditions), and the text to be labelled. However, we adopted the modification from \citet{hellwig2025we}, who additionally instructed the LLM to explicitly extract the exact aspect and opinion phrases from the input text. From the training split of the respective dataset and task, either 0, 10, or 50 few-shot examples were randomly sampled and inserted into the prompt, similar to \citet{hellwig2025we}.

Following prior work \citep{gou2023mvp,hellwig2025we,hu2022improving}, we generated multiple outputs with the LLM using five different random seeds. A predicted tuple was included in the final label set if it appeared in the majority (i.e., at least 3 out of 5) of the generated outputs. Formally:

\begin{equation}
T_{\text{final}} = \left\{ t \,\middle|\, \sum_{i=1}^{m} \mathbf{1}\left[t \in T_i\right] > \frac{m}{2} \right\}
\end{equation}

where $T_i$ is the set of predicted tuples from the $i$-th run, $m = 5$ is the number of generations, and $\mathbf{1}[\cdot]$ is the indicator function that returns 1 if the condition is true and 0 otherwise. 

Following \citet{hellwig2025we}, we re-executed the LLM in cases where it predicted an invalid aspect category or sentiment polarity, or generated phrases that were not present in the input sentence (see Algorithm \ref{alg:llm-fewshot-validation}). The label was set to an empty list (\texttt{[]}) if no valid tuples were predicted after 10 regenerations. To allow for sufficient variation across generations, the decoding temperature was set to 0.8, similar to \citet{hellwig2025we}.

Similar to \citet{hellwig2025we}, we employed Gemma-3-27B.\footnote{\url{https://ollama.com/library/gemma3:27b}} Gemma-3-27B is the most recent open-source LLM by \textit{Google}, which comprises 27.4 billion parameters \citep{team2025gemma}.

This and all subsequently introduced approaches were conducted on a single NVIDIA RTX A5000 GPU with 24GB VRAM, ensuring consistent computational environments across model training and evaluation procedures.

\subsubsection{Trainer Module}\label{sec:trainer-module}

Subsequently, the LLM-annotated examples were used as training data, combined with the given 0, 10, or 50 gold-standard examples. We evaluated two lightweight fine-tuning approaches based on \texttt{T5-base} (223 million parameters): (1) DLO \citep{hu2022improving} and (2) Paraphrase \citep{zhang2021aspect}. For both DLO and Paraphrase, we adopted the same hyperparameters as proposed by \citet{hu2022improving} and \citet{zhang2021aspect} respectively: 20 epochs, a learning rate of 2e-4 (3e-4 for Paraphrase) and a batch size of 16.


\subsection{Baselines}

\subsubsection{Fine-tuning on Human-Annotated Training Data}

We included fine-tuning baselines Paraphrase and DLO on varying amounts of human-annotated data. Specifically, we conduct experiments using 10, 50 or all examples from the training set for each task and dataset combination. We employed the same hyperparameters as specified in Section~\ref{sec:trainer-module} to ensure fair comparison.

\subsubsection{LLMs for Zero-shot and Few-shot Prompting}

Next, we incorporate the zero-shot and few-shot ICL results reported by \citet{hellwig2025we}, which demonstrated SOTA performance in low-resource scenarios for both TASD and ASQP. Following their experimental setup, we employed Gemma-3-27B for 0-, 10-, and 50-shot prompting configurations. 

\subsubsection{Low-resource Enhancement Methods.}

\begin{algorithm}[h]
\small
\DontPrintSemicolon
\SetAlgoLined
\SetKwFunction{GetSyn}{get\_synonym}
\SetKwFunction{Tok}{tokenizer}
\SetKwFunction{Insert}{insertion}
\SetKwFunction{Delete}{deletion}
\SetKwFunction{Swap}{swap}
\SetKwFunction{Replace}{synonym}
\SetKwFunction{Join}{join}

\KwIn{Dataset $\mathcal{D}$ with $k$ examples $(s, t)$, where $s$ is a sentence and $t$ the list of corresponding tuples; Number of augmentations per example $\alpha$}
\KwOut{Augmented dataset $\mathcal{D}_{\text{aug}}$ with $\alpha$ examples per $(s, t)$}

\ForEach{$(s, t) \in \mathcal{D}$}{
    $tokens \gets$ \Tok{$s$} \;

    \For{$i \gets 1$ \KwTo $\alpha$}{
        $tokens_{\text{aug}} \gets tokens.\text{copy()}$ \;
        $t_{\text{aug}} \gets t.\text{copy()}$ \;

        $tokens_{\text{aug}} \gets$ \Insert{$tokens_{\text{aug}}$} \;
        $tokens_{\text{aug}} \gets$ \Delete{$tokens_{\text{aug}}$} \;
        $tokens_{\text{aug}} \gets$ \Swap{$tokens_{\text{aug}}$} \;
        $tokens_{\text{aug}} \gets$ \Replace{$tokens_{\text{aug}}$} \;

        $s_{\text{aug}} \gets$ \Join{$tokens_{\text{aug}}$} \;
        $\mathcal{D}_{\text{aug}} \gets \mathcal{D}_{\text{aug}} \cup \{(s_{\text{aug}}, t_{\text{aug}})\}$ \;
    }
}
\Return $\mathcal{D}_{\text{aug}}$ \;
\caption{EDA}
\label{alg:eda-augmentation}
\end{algorithm}

\textbf{Easy Data Augmentation (EDA)} was evaluated since it's widely used across various NLP tasks and ABSA tasks considering less sentiment elements \citep{hsu2021semantics,liesting2021data,rahamim2023text}. For a given set of annotated examples (10 or 50), we generated 10 augmented examples. For 10 augmentations per example and 10 given examples, we produced 100 augmented examples, resulting in a total of 110 examples when combined with the given 10 real examples. 

The augmentation procedure is outlined in Algorithm \ref{alg:eda-augmentation}. First, the insertion operation adds a synonym of a randomly selected word from the sentence at a random position, but not within aspect or opinion terms. The deletion operation removes a random token that is not part of a term, swap exchanges two tokens that are not located within any term. Finally, the replacement operation substitutes a random token in the sentence with its synonym. DLO was employed for training.

\textbf{Quantity Augmentation and Information Enhancement (QAIE)} by \citet{lu2025qaie} was included with the following modifications: (1) augmentation was performed based on 10 or 50 given examples, consistent with the other experimental conditions, and (2) instead of \textit{Google}'s deprecated PaLM API\footnote{\textit{Google} Generative AI: \url{https://pypi.org/project/google-generativeai/}}, we used Gemma-3-27B, which further improves comparability across the evaluated approaches. In cases where an example did not contain any explicit aspect term, term-based augmentation was impossible, resulting in fewer augmentations compared to the ASQP task, where an opinion term is always given for all examples. Notably, \citet{lu2025qaie} exclusively augmented ASQP examples.

\textbf{DS\textsuperscript{2}-ABSA} was adapted with several modifications. To maintain consistency with LA-ABSA and the aforementioned baseline methods, we employed Gemma-3-27B instead of GPT-3.5-Turbo for data synthesis. Furthermore, we did not implement the iterative noisy self-training mechanism from the original DS\textsuperscript{2}-ABSA framework, where a teacher model iteratively refines synthetic labels and trains successive student models until validation performance stabilizes. Our scenario assumes strict data constraints: apart from 0, 10, or 50 training examples, no additional annotated data is assumed.



\subsection{Evaluation and Datasets}

\begin{table}[h]
\centering
\label{tab:dataset_stats}
\resizebox{0.9\columnwidth}{!}{%
\Large
\begin{tabular}{
>{\raggedright\arraybackslash}p{3cm}|*{2}{>{\centering\arraybackslash}p{1.4cm}}|*{2}{>{\centering\arraybackslash}p{1.4cm}}
}
\toprule
\multirow{2}{*}{\begin{tabular}[c]{@{}c@{}}\textbf{}\\\textbf{Dataset}\end{tabular}} & 
\multicolumn{2}{c|}{\textbf{\Large Train}} & 
\multicolumn{2}{c}{\textbf{\Large Test}} \\
\cmidrule(lr){2-3} \cmidrule(lr){4-5}
& \textbf{\large TASD} & \textbf{\large ASQP} & 
\textbf{\large TASD} & \textbf{\large ASQP} \\
\midrule
\rowcolor{gray!5}
\textbf{Rest15} & 1,120 & 834 & 582 & 537 \\
\textbf{Rest16} & 1,708 & 1,264 & 587 & 544 \\
\rowcolor{gray!5}
\textbf{FlightABSA} & 1,351 & 1,351 & 387 & 387 \\
\textbf{Coursera} & 1,400 & 1,400 & 400 & 400 \\
\rowcolor{gray!5}
\textbf{Hotels} & 1,400 & 1,400 & 400 & 400 \\
\bottomrule
\end{tabular}%
}
\caption{\textbf{Dataset statistics overview.} Distribution of annotated examples for TASD and ASQP tasks across five domain-specific datasets, showing training and test set sizes.}

\end{table}

\definecolor{lightgray}{gray}{0.6}

\begin{table*}[h]
\centering
\small
\setlength{\tabcolsep}{4pt}
\resizebox{2.0\columnwidth}{!}{%
\begin{tabular}{@{}llccccccccccc|cc@{}}
\toprule
\multirow{2}{*}{\textbf{Approach}} & \multicolumn{2}{c}{\textbf{\# Train}} & \multicolumn{2}{c}{\textbf{Rest15}} & \multicolumn{2}{c}{\textbf{Rest16}} & \multicolumn{2}{c}{\textbf{FlightABSA}} & \multicolumn{2}{c}{\textbf{Coursera}} & \multicolumn{2}{c}{\textbf{Hotels}} & \multicolumn{2}{c}{\textbf{AVG}} \\
\cmidrule(lr){2-3} \cmidrule(lr){4-5} \cmidrule(lr){6-7} \cmidrule(lr){8-9} \cmidrule(lr){10-11} \cmidrule(l){12-13} \cmidrule(l){14-15}
& \textbf{TASD} & \textbf{ASQP} & \textbf{TASD} & \textbf{ASQP} & \textbf{TASD} & \textbf{ASQP} & \textbf{TASD} & \textbf{ASQP} & \textbf{TASD} & \textbf{ASQP} & \textbf{TASD} & \textbf{ASQP} & \textbf{TASD} & \textbf{ASQP} \\
\arrayrulecolor{black}\midrule
\multicolumn{15}{@{}l}{\cellcolor{gray!15}\textbf{Scenario: 0 annotated examples given}} \\
\arrayrulecolor{black}\midrule
Gemma-3-27B (0-shot) & 0 & 0 & \textbf{30.36} & \textbf{24.73} & 45.51 & \textbf{28.96} & 51.81 & 42.37 & 29.50 & 13.36 & 38.97 & 23.02 & 39.23 & \textbf{26.49} \\
\noalign{\vskip 0.1cm}              
\arrayrulecolor{lightgray}\hline  
\noalign{\vskip 0.1cm}
\rowcolor{gray!5}
LA-ABSA &  &  &  &  &  &  &  &  &  &  &  &  &  &  \\
\textit{w/ Paraphrase} & full & full & 27.70 & 18.54 & \textbf{47.73} & 26.07 & \textbf{52.10} & 42.58 & \textbf{30.92} & \textbf{13.41} & \textbf{40.74} & \textbf{26.27} & \textbf{39.84} & 25.37 \\
\rowcolor{gray!5}
\textit{w/ DLO }& full & full & 25.82 & 18.88 & 47.41 & 26.97 & 51.75 & \textbf{42.82} & 30.60 & 13.38 & 40.70 & 24.97 & 39.26 & 25.40 \\
\arrayrulecolor{black}\midrule
\multicolumn{15}{@{}l}{\cellcolor{gray!15}\textbf{Scenario: 10 annotated examples given}} \\
\arrayrulecolor{black}\midrule
Gemma-3-27B ICL & 10 & 10 & \textbf{54.47} & \textbf{39.95} & \textbf{66.75} & \textbf{46.23} & 60.36 & 45.24 & \textbf{41.69} & 22.31 & \textbf{56.51} & 31.41 & \textbf{55.96} & \textbf{37.03} \\
\noalign{\vskip 0.1cm}              
\arrayrulecolor{lightgray}\hline  
\noalign{\vskip 0.1cm}
\rowcolor{gray!5}
Paraphrase & 10 & 10 & 8.75 & 1.32 & 6.66 & 3.56 & 8.82 & 3.44 & 15.94 & 4.75 & 14.91 & 2.63 & 11.02 & 3.14 \\
DLO & 10 & 10 & 15.84 & 4.37 & 13.59 & 5.18 & 16.07 & 4.87 & 22.93 & 4.47 & 18.07 & 3.53 & 17.30 & 4.48 \\
\noalign{\vskip 0.1cm}              
\arrayrulecolor{lightgray}\hline  
\noalign{\vskip 0.1cm}
\rowcolor{gray!5}
LA-ABSA &  &  &  &  &  &  &  &  &  &  &  &  &  &  \\
\textit{w/ Paraphrase} & full & full & 49.09 & 35.04 & 62.74 & 44.59 & 59.87 & 45.89 & 38.46 & 22.47 & 55.69 & \textbf{32.83} & 53.17 & 36.16 \\
\rowcolor{gray!5}
\textit{w/ DLO }& full & full & 49.23 & 37.19 & 62.37 & 46.20 & \textbf{61.40} & \textbf{46.47} & 39.22 & \textbf{23.37} & 55.27 & 31.44 & 53.50 & 36.93 \\
\noalign{\vskip 0.1cm}              
\arrayrulecolor{lightgray}\hline  
\noalign{\vskip 0.1cm}
EDA & 110 & 110 & 30.18 & 9.34 & 19.16 & 11.80 & 22.66 & 11.95 & 28.83 & 14.19 & 25.65 & 7.88 & 25.30 & 11.03 \\
\rowcolor{gray!5}
QAIE & 26.4 & 45.2 & 20.11 & 9.96 & 12.37 & 14.78 & 17.76 & 15.38 & 25.18 & 17.54 & 21.69 & 9.86 & 19.42 & 13.51 \\
DS\textsuperscript{2}-ABSA & 21,1k & 21,1k & 29.04 & 18.19 & 32.09 & 23.52 & 27.01 & 16.49 & 17.81 & 7.61 & 31.87 & 11.78 & 27.56 & 15.52 \\
\arrayrulecolor{black}\midrule
\multicolumn{15}{@{}l}{\cellcolor{gray!15}\textbf{Scenario: 50 annotated examples given}} \\
\arrayrulecolor{black}\midrule
\rowcolor{gray!5}
Gemma-3-27B ICL & 50 & 50 & \textbf{62.12} & \textbf{41.74} & \textbf{68.53} & \textbf{51.10} & \textbf{64.60} & 48.37 & \textbf{44.80} & \textbf{25.86} & \textbf{62.97} & 43.83 & \textbf{60.60} & \textbf{42.18} \\
\noalign{\vskip 0.1cm}              
\arrayrulecolor{lightgray}\hline  
\noalign{\vskip 0.1cm}
Paraphrase & 50 & 50 & 36.92 & 25.55 & 35.87 & 23.50 & 33.57 & 17.98 & 34.26 & 19.38 & 40.10 & 23.09 & 36.14 & 21.90 \\
\rowcolor{gray!5}
DLO & 50 & 50 & 39.54 & 26.63 & 43.95 & 29.57 & 42.92 & 28.74 & 36.04 & 19.08 & 44.72 & 27.20 & 41.44 & 26.24 \\
\noalign{\vskip 0.1cm}              
\arrayrulecolor{lightgray}\hline  
\noalign{\vskip 0.1cm}
LA-ABSA &  &  &  &  &  &  &  &  &  &  &  &  &  &  \\
\rowcolor{gray!5}
\textit{w/ Paraphrase} & full & full & 56.21 & 37.61 & 62.20 & 46.76 & 62.47 & 47.16 & 44.36 & 25.71 & 60.58 & 43.19 & 57.16 & 40.09 \\
\textit{w/ DLO }& full & full & 58.40 & 40.38 & 62.03 & 49.85 & 62.57 & \textbf{48.98} & 44.39 & 25.69 & 61.43 & \textbf{44.24} & 57.76 & 41.83 \\
\noalign{\vskip 0.1cm}              
\arrayrulecolor{lightgray}\hline  
\noalign{\vskip 0.1cm}
\rowcolor{gray!5}
EDA & 550 & 550 & 44.14 & 30.93 & 45.72 & 34.59 & 45.64 & 36.73 & 37.89 & 24.57 & 45.72 & 32.51 & 43.82 & 31.86 \\
QAIE & 141.8 & 248.8 & 45.01 & 33.87 & 45.09 & 35.21 & 48.44 & 33.98 & 36.23 & 22.45 & 50.51 & 35.95 & 45.06 & 32.29 \\
\rowcolor{gray!5}
DS\textsuperscript{2}-ABSA & 21,7k & 21,6k & 37.94 & 28.93 & 43.53 & 35.79 & 33.39 & 23.87 & 29.56 & 16.41 & 40.26 & 23.61 & 36.94 & 25.72 \\
\arrayrulecolor{black}\midrule
\multicolumn{15}{@{}l}{\cellcolor{gray!15}\textbf{Full set of human-annotated examples: SOTA approaches}} \\
\arrayrulecolor{black}\midrule
Paraphrase & full & full & \textbf{\underline{63.06}} & 46.93 & \textbf{\underline{71.97}} & 57.93 & \textbf{\underline{69.74}} & 57.76 & 51.86 & 32.34 & 67.70 & 53.87 & 64.87 & 49.77 \\
\rowcolor{gray!5}
DLO & full & full & 62.95 & \textbf{\underline{48.18}} & 71.79 & \textbf{\underline{59.79}} & 68.95 & \textbf{\underline{58.33}} & \textbf{\underline{52.58}} & \textbf{\underline{32.54}} & \textbf{\underline{68.56}} & \textbf{\underline{55.45}} & \textbf{\underline{64.97}} & \textbf{\underline{50.86}} \\
\bottomrule
\end{tabular}
}
\caption{\textbf{F1 scores of LA-ABSA}. Results are evaluated against EDA-based data augmentation methods, QAIE (FT), DS\textsuperscript{2}-ABSA (FT), and prompting baselines (0, 10, and 50 annotated examples) as reported by \citet{hellwig2025we}, as well as fully supervised models including DLO \citep{hu2022improving} (FT) and Paraphrase \citep{zhang2021aspect} (FT). The highest F1 scores within each annotation regime (0, 10, 50 or all examples) are shown in bold; the best overall scores across all settings are underlined.}

\label{tab:results}
\end{table*}

We evaluated the tuple prediction tasks ASQP and TASD on five diverse datasets, which, to the best of our knowledge, constitute the only publicly available resources annotated for both tasks. For the TASD task, we utilized the SemEval 2015 \citep{pontiki2015semeval} and 2016 \citep{pontiki2016semeval} datasets. For ASQP, we employed the extended versions of the SemEval datasets by \citet{zhang2021aspect}, which additionally include opinion term annotations.

Furthermore, we incorporated an adapted version of the OATS dataset \citep{chebolu2024oats} by \citet{hellwig2025we}, which comprises hotel and e-learning (Coursera) domain data. These adaptations included converting ASQP quadruples to TASD triplets by removing opinion terms and eliminating duplicate triplets from the resulting dataset. Finally, we included FlightABSA, an ABSA-dataset comprising airline reviews that was introduced by \citet{hellwig2025we}. The distribution of aspect categories and polarities across the considered datasets and tasks is illustrated in Appendix \ref{sec:distributions}. 

As commonly done in ABSA research, the reported evaluation metric is the micro-averaged F1 score \citep{zhang2022survey}.

%% file: 04_results.tex
\section{Results}

\begin{figure*}[h]
    \centering
    \includegraphics[width=\textwidth]{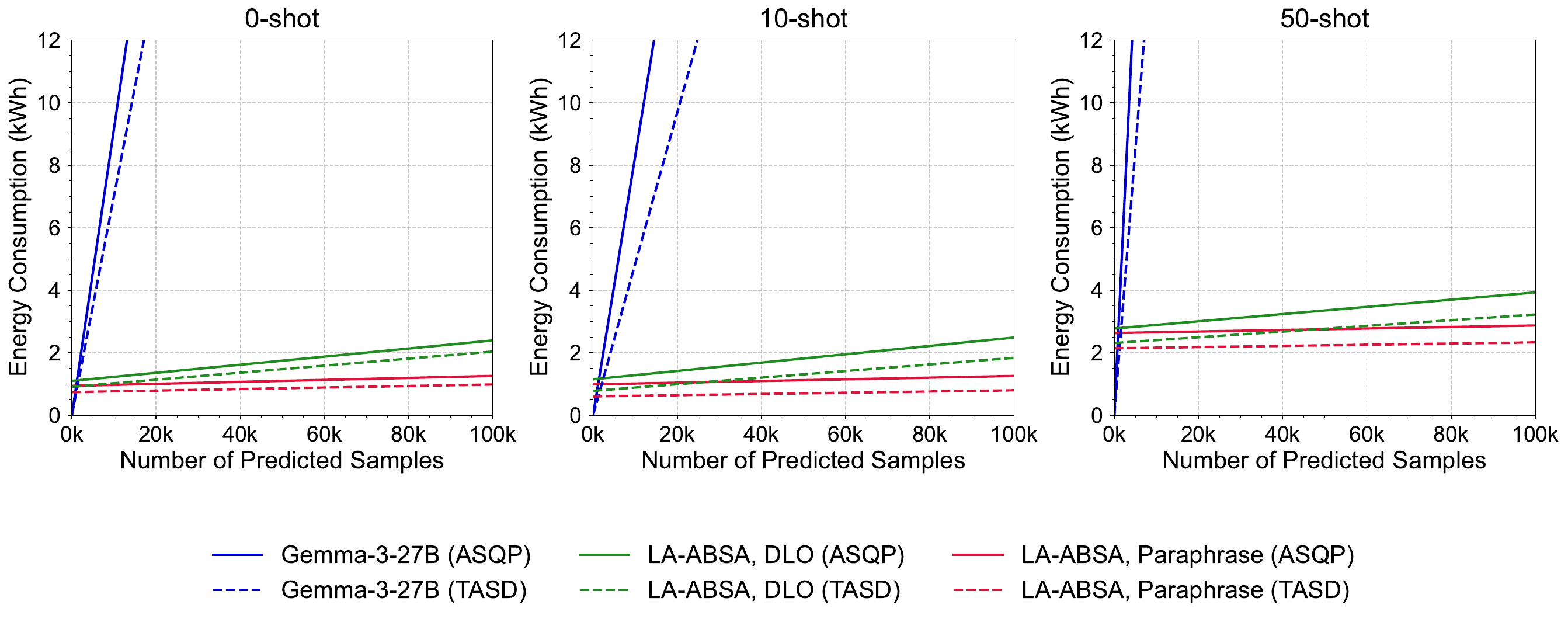}
    \caption{\textbf{Energy consumption analysis}. Comparative evaluation of energy usage (kilowatt-hours, kWh) for ASQP and TASD tasks across different settings (0, 10 or 50 annotated examples given). Results are shown for LLM-prompting (Gemma-3-27B) and LA-ABSA methods, using either DLO \citep{hu2022improving} or Paraphrase \citep{zhang2021aspect} for fine-tuning. Each line represents the average energy usage to predict up to 100,000 examples per method and task across the five datasets. LA-ABSA approaches generally require much lower energy due to their smaller underlying base model T5-base.}
    \label{fig:energy}
\end{figure*}

\subsection{Overall Results}

Table \ref{tab:results} presents the performance scores of LA-ABSA. Detailed precision, recall, and macro-averaged F1 scores are reported in Appendix \ref{appendix:pre-rec-performance}.

\textbf{LA-ABSA achieves competitive or superior performance relative to LLM prompting.} Table \ref{tab:results} shows that LA-ABSA achieved F1 scores only about 4 percentage points below those achieved by LLM-prompting (e.g., Rest15-TASD, 50-shot: 58.40 vs. 62.12). However, several notable exceptions emerge: in zero-shot scenarios, LA-ABSA with DLO fine-tuning outperformed zero-shot LLM-prompting across most datasets and tasks. Additionally, occasional superior performance is observed in 10-shot and 50-shot settings for the ASQP task.

\textbf{LA-ABSA demonstrates higher performance scores than data augmentation approaches.} Augmentation approaches that expand labelled examples, including EDA-based methods, QAIE and DS\textsuperscript{2}-ABSA demonstrated substantially lower performance than LA-ABSA across 0-shot, 10-shot, and 50-shot learning scenarios. Performance gaps reach approximately 10 percentage points in several cases, with differences exceeding 20 percentage points in 10-shot learning configurations. 

To investigate the impact of different augmentation ratios (2, 5, 10, and 15 augmentations per annotated example) and fine-tuning approaches (DLO, Paraphrase), we conducted an ablation study for EDA detailed in Appendix \ref{appendix:eda-ablation}. Our results demonstrate that 10 augmentations in combination with DLO achieve the highest performance. 

\textbf{Expert-annotated training data yielded superior performance over LLM-generated annotations.} Fine-tuning SOTA approaches for tuple prediction tasks using expert annotations is still superior to using LLM-generated annotations. Furthermore, fine-tuning on expert annotations still surpasses all other evaluated approaches in our experimental setup. 

Notably, we conducted two-sided paired t-tests with Bonferroni-Holm correction \citep{Holm1979-zm} to examine significant differences ($p_{\text{adj}} \leq 0.05$) between LA-ABSA (based on Paraphrase or DLO), fine-tuned approaches trained on the full human-labelled training set (DLO and Paraphrase), and the strongest baseline in low-resource scenarios, ICL with Gemma-3-27B. Considering these five comparison groups, we performed 10 pairwise tests per task (TASD and ASQP) and shot setting (0, 10, and 50 shots), yielding a total of 56 comparisons. Each group comprised five values (one per dataset). No significant differences were found, except that Paraphrase and DLO fine-tuned on the full set achieved significantly higher performance scores than 0-shot prompting for the TASD task.

\begin{table*}[h]
\centering
\resizebox{2.0\columnwidth}{!}{%
\begin{tabular}{@{}lccccccc@{}}
\toprule
\multirow{2}{*}{\textbf{Method}} & \multicolumn{2}{c}{\textbf{0-shot}} & \multicolumn{2}{c}{\textbf{10-shot}} & \multicolumn{2}{c}{\textbf{50-shot}} \\
\cmidrule(lr){2-3} \cmidrule(lr){4-5} \cmidrule(lr){6-7}
 & \textbf{TASD} & \textbf{ASQP} & \textbf{TASD} & \textbf{ASQP} & \textbf{TASD} & \textbf{ASQP} \\
\midrule
\rowcolor{gray!5} LA-ABSA (Paraphrase) & 2.47 ± 0.53 & 3.17 ± 0.31 & 1.96 ± 0.40 & 2.66 ± 0.28 & 1.83 ± 0.38 & 2.41 ± 0.46 \\
LA-ABSA (DLO) & 11.16 ± 0.72 & 13.06 ± 0.89 & 10.68 ± 2.11 & 13.19 ± 1.62 & 8.95 ± 0.85 & 11.65 ± 2.23 \\
\rowcolor{gray!5} Gemma-3-27B ICL & 703.10 ± 63.97 & 912.25 ± 72.67 & 481.39 ± 46.11 & 843.68 ± 356.29 & 1688.76 ± 255.30 & 2963.65 ± 1360.88 \\
\bottomrule
\end{tabular}
}
\caption{\textbf{Average energy consumption analysis}. Energy usage (milliwatt-hours, mWh) per sample of LA-ABSA and LLM-prompting (Gemma-3-27B) across few-shot settings. Results show mean ± standard deviation across the five datasets for TASD and ASQP tasks. We performed 18 pairwise comparisons between DLO, Paraphrase, and Prompting (3 comparisons $\times$ 2 tasks $\times$ 3 shot settings) using paired t-tests with Bonferroni-Holm correction ($p_{\text{adj}} \leq 0.05$), and all differences were found to be significant.}
\label{tab:energy_consumption}

\end{table*}

\subsection{Energy Consumption Analysis}

A key advantage of LA-ABSA lies in its superior energy efficiency compared to LLM-prompting approaches. To quantify this benefit, we conducted an energy consumption analysis comparing the two best-performing low-resource methods LA-ABSA and LLM-prompting across up to 100,000 generated examples.

Our analysis uses the average energy consumption per prediction as the baseline metric, computed across all five datasets. It is important to note that LA-ABSA's energy consumption profile includes initial overhead costs from two prerequisite steps: (1) LLM-based annotation of training examples, and (2) fine-tuning of the lightweight approach (DLO or Paraphrase). Consequently, the energy curve for LA-ABSA does not originate at zero, reflecting these upfront investments. 

As illustrated in Figure \ref{fig:energy}, LA-ABSA demonstrates significantly lower energy consumption than LLM-prompting when processing fewer than 2,000 examples, regardless of the task type or the number of given examples (0, 10, or 50). This finding indicates that despite the initial energy investment required for model preparation, LA-ABSA becomes substantially more energy-efficient at scale, positioning it as a more sustainable solution for large-scale ABSA deployments.

Table \ref{tab:energy_consumption} provides detailed energy consumption metrics for LA-ABSA, expressed in milliwatt-hours (mWh) per predicted example. The results highlight the particular efficiency of the paraphrase approach, which evaluates only a single ordering of sentiment elements per test example, compared to the DLO approach that requires assessment of multiple sentiment element permutations.

We note that all energy measurements were conducted using an NVIDIA A5000 GPU (24GB VRAM), and the reported metrics are inherently hardware-dependent. Generalization to other GPU architectures would require appropriate scaling factors. Additionally, we provide a complementary analysis of prediction times, which exhibits trends consistent with our energy consumption findings (see Appendix \ref{sec:time}).

%% file: 05_discussion.tex
\section{Discussion and Limitations}

In this section, we discuss the key findings and potential limitations of our study.

First, our proposed approach LA-ABSA demonstrated superior performance compared to previous approaches that rely on only a few annotated examples and utilize smaller models (e.g., T5) rather than LLMs, which is an observation  also made in other NLP tasks \citep{li2025learning, huang2022large}. Data augmentation approaches such as Easy Data Augmentation (EDA) \citep{liesting2021data}, QAIE \citep{lu2025qaie} and DS\textsuperscript{2}-ABSA \citep{xu2025ds2absa} were adapted for TASD and ASQP tasks for the first time in this study, as previous work only applied these augmentation strategies to ABSA tasks with fewer sentiment elements. Nevertheless, LA-ABSA demonstrated much higher F1 scores across scenarios and datasets.

In contrast, LLM-based ICL marginally outperforms LA-ABSA in most cases, with few exceptions. Figure \ref{fig:energy} demonstrates that LA-ABSA achieves substantially higher energy efficiency when processing more than 2,000 examples. Given the magnitude of this difference (e.g., for the TASD task, LA-ABSA + Paraphrase requires 2.47 mWh/example vs. Gemma-3-27B at 703.1 mWh/example), similar efficiency gains would likely be observed even when comparing ICL with smaller LLMs (e.g., 12B parameters) or single-execution approaches without self-consistency learning (majority voting mechanism).

In this context, a limitation of this study is the evaluation of only one LLM, primarily due to the extensive computational requirements of the investigated conditions. Examining reasoning models that generate intermediate reasoning steps, which have demonstrated superiority over non-reasoning approaches across various NLP tasks \citep{wei2022chain, fei2023reasoning, sun2023text} might be a strategy to further improve the annotation quality. However, such models substantially increase inference time due to the higher number of output tokens generated. 

Finally, LA-ABSA did not reach performance levels comparable to models fine-tuned on human-annotated data. On the ASQP task, LA-ABSA demonstrated an average performance decrease of 9 percentage points relative to the approach fine-tuned on human annotations.



%% file: 06_conclusion.tex
\section{Conclusion \& Future Work}

In this work, we introduced LA-ABSA, a novel approach that leverages LLM-generated annotations to fine-tune lightweight models for ABSA tasks. Our comprehensive evaluation across five datasets and two tuple prediction tasks (TASD and ASQP) demonstrated competitive performance with LLM-based in-context learning approaches while offering substantial energy efficiency advantages.


Future work could explore several promising directions. First, crowd-sourcing approaches could be explored as an alternative or complement to LLM-based annotation, potentially offering a middle ground between fully automated and expert annotation. Second, data synthesis approaches present an interesting avenue for investigation. While this study assumed the availability of non-annotated texts, low-resource scenarios may arise where such texts are unavailable. In such cases, similar to approaches practiced in other NLP tasks or ABSA tasks, methods could be developed to synthetically generate opinion pieces (e.g., reviews) using LLMs \citep{xu2025ds2absa,ye2022zerogen,yu2023large}, as well as approaches that generate annotated training examples in a single step \citep{xu2025ds2absa, hellwig2025exploring}. 

%% file: 07_ethical_considerations.tex
\section{Ethics Statement}

This research was conducted without industrial funding or commercial sponsorship. All datasets contain annotated examples that have not been anonymized, except for FlightABSA. To ensure research transparency and reproducibility, code is available on GitHub. Claude Sonnet 4.5\footnote{Claude Sonnet: \url{https://www.anthropic.com/claude/sonnet}} was used to assist in the formulation of this publication. 

Finally, we want to highlight that LA-ABSA contributes to the reduction of the carbon footprint associated with large-scale model inference. By utilizing a substantially smaller base model and requiring fewer computational resources, LA-ABSA minimizes energy consumption. This resource-efficient design supports more sustainable research practices and aligns with broader efforts to mitigate the environmental impact of AI systems.

%% file: 08_appendix.tex
\onecolumn
\appendix

\section{Datasets: Aspect Category and Sentiment Distributions}\label{sec:distributions}

\begin{figure*}[h!]
    \centering
    \includegraphics[width=\textwidth]{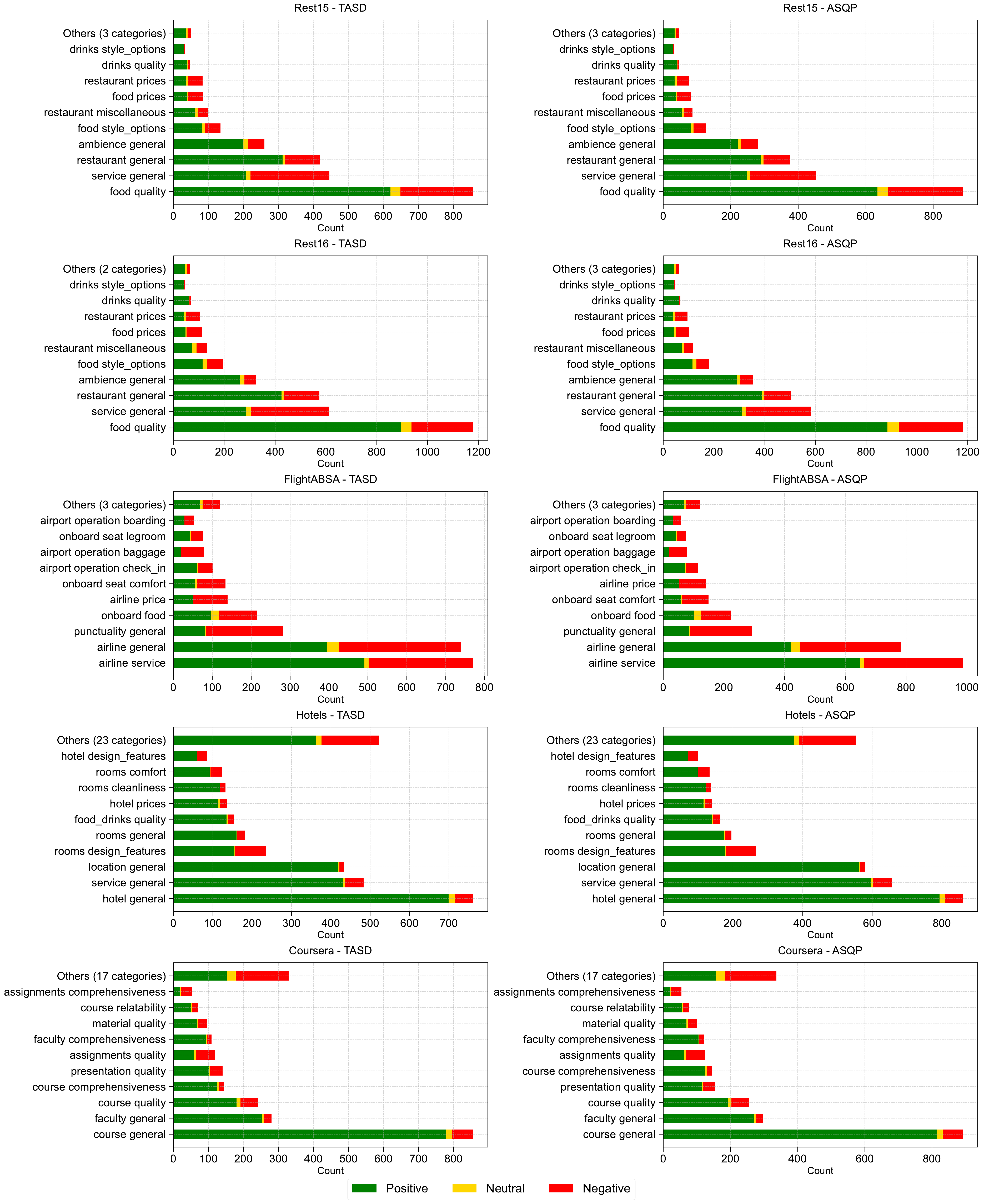}
    \caption{\textbf{Distribution of aspect categories and sentiments across datasets for TASD and ASQP tasks.} Each subplot shows the top 10 aspect categories (sorted by total frequency), with stacked bars representing positive (green), neutral (yellow), and negative (red) sentiments. The 'Others' category aggregates the remaining aspects. Results are aggregated across five datasets: Rest15 \citep{zhang2021aspect, pontiki2015semeval}, Rest16 \citep{zhang2021aspect, pontiki2016semeval}, FlightABSA \citep{hellwig2025we}, Coursera \citep{chebolu2024oats}, and Hotels \citep{chebolu2024oats}. This visualization highlights the imbalances in aspect-level sentiment annotations, showing varying distributions of polarities and aspect categories across datasets.}
    \label{fig:ac_pol_dist}
\end{figure*}

\clearpage

\section{Ablation Study: EDA}\label{appendix:eda-ablation}

\begin{table}[h]
\centering
\small
\setlength{\tabcolsep}{4pt}
\resizebox{1.0\columnwidth}{!}{%
\begin{tabular}{@{}l@{\hspace{4pt}}c@{\hspace{8pt}}cc@{\hspace{12pt}}cc@{\hspace{12pt}}cc@{\hspace{12pt}}cc@{\hspace{12pt}}cc@{\hspace{12pt}}cc@{\hspace{8pt}}|@{\hspace{8pt}}cc@{}}
\toprule
\multirow{2}{*}{\textbf{Approach}} & \multirow{2}{*}{\textbf{\# Aug}} & \multicolumn{2}{c@{\hspace{12pt}}}{\textbf{\# Train}} & \multicolumn{2}{c@{\hspace{12pt}}}{\textbf{Rest15}} & \multicolumn{2}{c@{\hspace{12pt}}}{\textbf{Rest16}} & \multicolumn{2}{c@{\hspace{12pt}}}{\textbf{FlightABSA}} & \multicolumn{2}{c@{\hspace{12pt}}}{\textbf{Coursera}} & \multicolumn{2}{c@{\hspace{8pt}}|@{\hspace{8pt}}}{\textbf{Hotels}} & \multicolumn{2}{c@{}}{\textbf{AVG}} \\
\cmidrule(lr){3-4} \cmidrule(lr){5-6} \cmidrule(lr){7-8} \cmidrule(lr){9-10} \cmidrule(lr){11-12} \cmidrule(lr){13-14} \cmidrule(l){15-16}
& & \small TASD & \small ASQP & \small TASD & \small ASQP & \small TASD & \small ASQP & \small TASD & \small ASQP & \small TASD & \small ASQP & \small TASD & \small ASQP & \small TASD & \small ASQP \\
\midrule
\multicolumn{16}{@{}l@{}}{\textbf{Scenario: 10 annotated examples given}} \\
\midrule
\addlinespace[2pt]
EDA w/ Paraphrase & 2 & 30 & 30 & 13.15 & 1.74 & 7.67 & 11.41 & 10.79 & 4.42 & 17.97 & 6.61 & 22.05 & 3.47 & 14.33 & 5.53 \\
\rowcolor{gray!5}
EDA w/ Paraphrase & 5 & 60 & 60 & 23.20 & 10.42 & 16.35 & 12.30 & 19.26 & 10.99 & 23.62 & 9.47 & 25.85 & 6.15 & 21.66 & 9.86 \\
EDA w/ Paraphrase & 10 & 110 & 110 & 26.07 & \textbf{10.52} & 16.82 & \textbf{12.68} & 21.83 & \textbf{13.46} & 25.67 & 10.35 & 24.95 & 5.84 & 23.07 & 10.57 \\
\rowcolor{gray!5}
EDA w/ Paraphrase & 15 & 160 & 160 & 25.53 & 10.08 & 18.34 & 12.32 & 21.57 & 12.39 & 25.60 & 12.33 & 24.94 & 7.03 & 23.20 & 10.83 \\
\addlinespace[2pt]
EDA w/ DLO & 2 & 30 & 30 & 24.58 & 6.74 & 16.97 & 9.27 & 21.18 & 12.13 & 28.14 & 9.75 & 22.73 & 5.98 & 22.72 & 8.77 \\
\rowcolor{gray!5}
EDA w/ DLO & 5 & 60 & 60 & 26.65 & 8.68 & 18.06 & 10.70 & 21.66 & 12.40 & 28.25 & 11.73 & 23.95 & 7.51 & 23.71 & 10.20 \\
EDA w/ DLO & 10 & 110 & 110 & \textbf{30.18} & 9.34 & 19.16 & 11.80 & \textbf{22.66} & 11.95 & \textbf{28.83} & \textbf{14.19} & 25.65 & \textbf{7.88} & \textbf{25.30} & \textbf{11.03} \\
\rowcolor{gray!5}
EDA w/ DLO & 15 & 160 & 160 & 29.10 & 8.07 & \textbf{19.59} & 11.83 & 22.05 & 12.13 & 27.40 & 13.36 & \textbf{27.04} & 6.94 & 25.04 & 10.46 \\
\midrule
\multicolumn{16}{@{}l@{}}{\textbf{Scenario: 50 annotated examples given}} \\
\midrule
\addlinespace[2pt]
EDA w/ Paraphrase & 2 & 150 & 150 & 42.70 & 27.55 & 41.59 & 28.78 & 40.73 & 28.26 & 36.47 & 19.90 & 43.44 & 27.96 & 40.98 & 26.49 \\
\rowcolor{gray!5}
EDA w/ Paraphrase & 5 & 300 & 300 & 42.29 & 28.96 & 42.57 & 29.40 & 42.80 & 30.32 & 36.85 & 21.08 & 44.67 & 28.04 & 41.84 & 27.56 \\
EDA w/ Paraphrase & 10 & 550 & 550 & 41.83 & 28.70 & 43.01 & 30.20 & 41.62 & 31.04 & 35.84 & 21.37 & 42.70 & 27.97 & 41.00 & 27.85 \\
\rowcolor{gray!5}
EDA w/ Paraphrase & 15 & 800 & 800 & 40.78 & 27.81 & 40.81 & 28.80 & 42.02 & 31.92 & 36.49 & 21.22 & 43.45 & 27.38 & 40.71 & 27.43 \\
\addlinespace[2pt]
EDA w/ DLO & 2 & 150 & 150 & 43.86 & 30.41 & 45.77 & 33.52 & 44.11 & 34.84 & 37.81 & 23.53 & \textbf{45.89} & 31.39 & 43.49 & 30.74 \\
\rowcolor{gray!5}
EDA w/ DLO & 5 & 300 & 300 & 43.95 & 30.90 & \textbf{46.28} & 34.42 & 44.90 & 36.37 & \textbf{38.32} & 23.53 & 45.20 & \textbf{32.55} & 43.73 & 31.55 \\
EDA w/ DLO & 10 & 550 & 550 & \textbf{44.14} & \textbf{30.93} & 45.72 & \textbf{34.59} & \textbf{45.64} & \textbf{36.73} & 37.89 & \textbf{24.57} & 45.72 & 32.51 & \textbf{43.82} & \textbf{31.86} \\
\rowcolor{gray!5}
EDA w/ DLO & 15 & 800 & 800 & 41.71 & 28.82 & 42.10 & 31.68 & 45.57 & 34.09 & 36.55 & 23.33 & 44.61 & 31.13 & 42.11 & 29.81 \\
\bottomrule
\end{tabular}
}
\caption{\textbf{Results of ablation study for EDA-based data augmentation.} F1 scores are reported across different augmentation ratios (2, 5, 10, 15) and fine-tuning approaches (DLO, Paraphrase) for scenarios with 10 and 50 annotated examples. Results demonstrate that 10 augmentations combined with DLO achieves optimal performance across most datasets. Bold values indicate the highest F1 scores within a scenario assuming either 10 or 50 given annotated examples.}

\label{tab:results-ablation-eda}
\end{table}

\section{LA-ABSA: Efficiency in Terms of Time}\label{sec:time}

\begin{figure*}[h!]
    \centering
    \includegraphics[width=\textwidth]{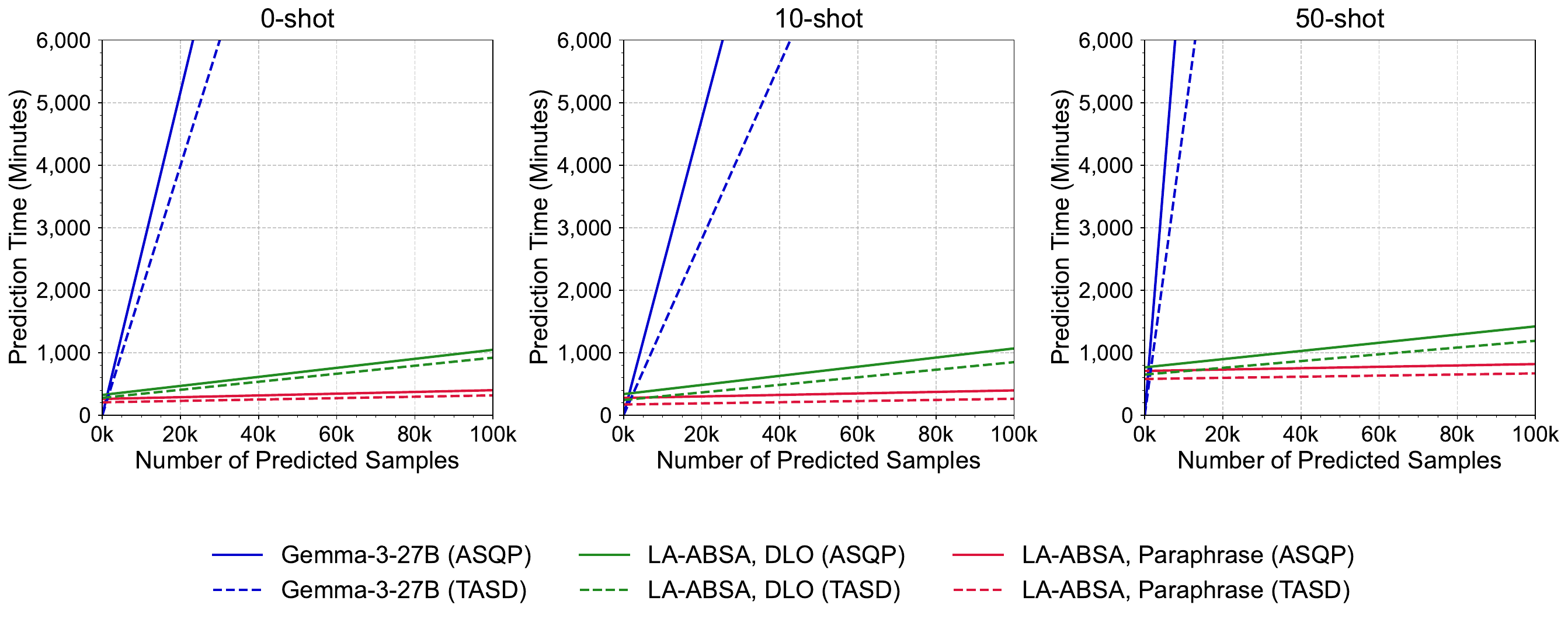}
    \caption{\textbf{Prediction time analysis}.  Prediction time in minutes for ASQP and TASD tasks across different settings (0, 10, or 50 annotated examples given). Results are shown for LLM-prompting (Gemma-3-27B) and LA-ABSA using DLO \citep{hu2022improving} and Paraphrase \citep{zhang2021aspect} for fine-tuning. Each line represents the average time required to predict up to 100,000 examples per method and task across the five datasets. LA-ABSA approaches generally require substantially less time due to their smaller underlying base model T5-base.}
    \label{fig:time}
\end{figure*}

\clearpage

\section{Performance: Precision, Recall and F1 Macro}\label{appendix:pre-rec-performance}

\subsection{Precision}

\begin{table}[h]
\setlength{\tabcolsep}{4pt}
\resizebox{\textwidth}{!}{%
\begin{tabular}{@{}llccccccccccc|cc@{}}
\toprule
\multirow{2}{*}{\textbf{Approach}} & \multicolumn{2}{c}{\textbf{\# Train}} & \multicolumn{2}{c}{\textbf{Rest15}} & \multicolumn{2}{c}{\textbf{Rest16}} & \multicolumn{2}{c}{\textbf{FlightABSA}} & \multicolumn{2}{c}{\textbf{Coursera}} & \multicolumn{2}{c}{\textbf{Hotels}} & \multicolumn{2}{c}{\textbf{AVG}} \\
\cmidrule(lr){2-3} \cmidrule(lr){4-5} \cmidrule(lr){6-7} \cmidrule(lr){8-9} \cmidrule(lr){10-11} \cmidrule(l){12-13} \cmidrule(l){14-15}
& \textbf{TASD} & \textbf{ASQP} & \textbf{TASD} & \textbf{ASQP} & \textbf{TASD} & \textbf{ASQP} & \textbf{TASD} & \textbf{ASQP} & \textbf{TASD} & \textbf{ASQP} & \textbf{TASD} & \textbf{ASQP} & \textbf{TASD} & \textbf{ASQP} \\
\arrayrulecolor{black}\midrule
\multicolumn{15}{@{}l}{\cellcolor{gray!15}\textbf{Scenario: 0 annotated examples given}} \\
\arrayrulecolor{black}\midrule
Gemma-3-27B (0-shot) & 0 & 0 & \textbf{29.41} & \textbf{23.35} & 44.49 & \textbf{27.75} & 47.55 & \textbf{39.70} & 26.95 & \textbf{11.95} & 37.12 & 22.88 & 37.11 & \textbf{25.13} \\
\noalign{\vskip 0.1cm}              
\arrayrulecolor{lightgray}\hline  
\noalign{\vskip 0.1cm}
\rowcolor{gray!5}
LA-ABSA &  &  &  &  &  &  &  &  &  &  &  &  &  &  \\
\textit{w/ Paraphrase} & full & full & 26.91 & 17.09 & \textbf{45.93} & 24.00 & \textbf{48.95} & 39.41 & \textbf{28.32} & 11.68 & \textbf{38.53} & \textbf{25.41} & \textbf{37.73} & 23.52 \\
\rowcolor{gray!5}
\textit{w/ DLO }& full & full & 24.44 & 17.21 & 45.68 & 24.62 & 47.41 & 39.06 & 27.52 & 11.48 & 37.98 & 23.83 & 36.61 & 23.24 \\
\arrayrulecolor{black}\midrule
\multicolumn{15}{@{}l}{\cellcolor{gray!15}\textbf{Scenario: 10 annotated examples given}} \\
\arrayrulecolor{black}\midrule
Gemma-3-27B ICL & 10 & 10 & \textbf{56.40} & \textbf{39.41} & \textbf{68.38} & \textbf{44.64} & 59.85 & 45.39 & \textbf{43.11} & \textbf{23.41} & \textbf{57.93} & \textbf{35.29} & \textbf{57.13} & \textbf{37.63} \\
\noalign{\vskip 0.1cm}              
\arrayrulecolor{lightgray}\hline  
\noalign{\vskip 0.1cm}
\rowcolor{gray!5}
Paraphrase & 10 & 10 & 10.72 & 1.64 & 7.61 & 4.02 & 10.44 & 4.34 & 17.69 & 5.35 & 18.74 & 3.66 & 13.04 & 3.80 \\
DLO & 10 & 10 & 19.23 & 4.64 & 13.27 & 5.49 & 19.02 & 6.15 & 25.45 & 5.03 & 18.84 & 3.68 & 19.16 & 5.00 \\
\noalign{\vskip 0.1cm}              
\arrayrulecolor{lightgray}\hline  
\noalign{\vskip 0.1cm}
\rowcolor{gray!5}
LA-ABSA &  &  &  &  &  &  &  &  &  &  &  &  &  &  \\
\textit{w/ Paraphrase} & full & full & 51.41 & 33.75 & 63.75 & 43.35 & 61.12 & \textbf{45.85} & 38.28 & 21.98 & 57.70 & 34.80 & 54.45 & 35.95 \\
\rowcolor{gray!5}
\textit{w/ DLO }& full & full & 50.33 & 35.54 & 63.51 & 43.85 & \textbf{61.29} & 45.63 & 38.82 & 22.79 & 56.07 & 33.10 & 54.00 & 36.18 \\
\noalign{\vskip 0.1cm}              
\arrayrulecolor{lightgray}\hline  
\noalign{\vskip 0.1cm}
EDA & 110 & 110 & 32.49 & 9.16 & 19.32 & 11.92 & 26.82 & 14.10 & 31.08 & 15.52 & 26.15 & 8.61 & 27.17 & 11.86 \\
\rowcolor{gray!5}
QAIE & 26.4 & 45.2 & 26.44 & 12.71 & 15.71 & 17.42 & 22.37 & 20.19 & 28.49 & 19.70 & 29.39 & 12.40 & 24.48 & 16.49 \\
DS\textsuperscript{2}-ABSA & 21,1k & 21,1k & 29.55 & 18.10 & 33.30 & 22.72 & 27.34 & 17.33 & 17.40 & 7.43 & 32.17 & 12.03 & 27.95 & 15.52 \\
\arrayrulecolor{black}\midrule
\multicolumn{15}{@{}l}{\cellcolor{gray!15}\textbf{Scenario: 50 annotated examples given}} \\
\arrayrulecolor{black}\midrule
\rowcolor{gray!5}
Gemma-3-27B ICL & 50 & 50 & \textbf{\underline{68.03}} & \textbf{44.57} & \textbf{\underline{71.52}} & \textbf{54.55} & \textbf{66.33} & \textbf{51.95} & \textbf{51.32} & \textbf{31.96} & \textbf{\underline{70.47}} & \textbf{53.39} & \textbf{\underline{65.54}} & \textbf{47.28} \\
\noalign{\vskip 0.1cm}              
\arrayrulecolor{lightgray}\hline  
\noalign{\vskip 0.1cm}
Paraphrase & 50 & 50 & 39.57 & 24.58 & 37.18 & 23.75 & 36.10 & 18.58 & 37.64 & 20.72 & 45.21 & 23.67 & 39.14 & 22.26 \\
\rowcolor{gray!5}
DLO & 50 & 50 & 40.48 & 24.92 & 44.59 & 29.09 & 42.01 & 28.30 & 39.26 & 20.44 & 49.18 & 28.54 & 43.10 & 26.26 \\
\noalign{\vskip 0.1cm}              
\arrayrulecolor{lightgray}\hline  
\noalign{\vskip 0.1cm}
LA-ABSA &  &  &  &  &  &  &  &  &  &  &  &  &  &  \\
\rowcolor{gray!5}
\textit{w/ Paraphrase} & full & full & 60.49 & 38.23 & 63.32 & 47.39 & 64.28 & 48.06 & 46.39 & 27.13 & 65.43 & 45.94 & 59.98 & 41.35 \\
\textit{w/ DLO }& full & full & 62.07 & 40.23 & 63.47 & 49.64 & 62.88 & 49.02 & 46.50 & 27.01 & 65.58 & 46.61 & 60.10 & 42.50 \\
\noalign{\vskip 0.1cm}              
\arrayrulecolor{lightgray}\hline  
\noalign{\vskip 0.1cm}
\rowcolor{gray!5}
EDA & 550 & 550 & 45.93 & 31.12 & 46.84 & 34.87 & 46.81 & 38.17 & 39.96 & 26.28 & 49.55 & 33.48 & 45.82 & 32.78 \\
QAIE & 141.8 & 248.8 & 48.89 & 35.07 & 48.29 & 37.15 & 51.71 & 37.13 & 37.70 & 23.57 & 58.26 & 38.78 & 48.97 & 34.34 \\
\rowcolor{gray!5}
DS\textsuperscript{2}-ABSA & 21,7k & 21,6k & 37.80 & 27.92 & 43.13 & 34.96 & 32.25 & 23.58 & 29.42 & 16.03 & 42.16 & 24.01 & 36.95 & 25.30 \\
\arrayrulecolor{black}\midrule
\multicolumn{15}{@{}l}{\cellcolor{gray!15}\textbf{Full set of human-annotated examples: SOTA approaches}} \\
\arrayrulecolor{black}\midrule
Paraphrase & full & full & - & 46.16 & - & 56.63 & \textbf{\underline{70.22}} & \textbf{\underline{57.37}} & 52.73 & \textbf{\underline{32.06}} & \textbf{68.41} & 52.61 & \textbf{63.79} & 48.97 \\
\rowcolor{gray!5}
DLO & full & full & - & \textbf{\underline{47.08}} & - & \textbf{\underline{57.92}} & 68.60 & 56.67 & \textbf{\underline{52.79}} & 32.03 & \textbf{68.41} & \textbf{\underline{54.39}} & 63.27 & \textbf{\underline{49.62}} \\
\bottomrule
\end{tabular}
}
\caption{\textbf{Precision scores of LA-ABSA}. Results are evaluated against EDA-based data augmentation methods, QAIE, DS\textsuperscript{2}-ABSA, and prompting baselines (0, 10, and 50 annotated examples) as reported by \citet{hellwig2025we}, as well as fully supervised models including DLO \citep{hu2022improving} and Paraphrase \citep{zhang2021aspect}. The highest precision scores within each annotation regime (0, 10, 50 or all examples) are shown in bold; the best overall scores across all settings are underlined.}

\label{tab:results-precision}
\end{table}

\clearpage
\subsection{Recall}
\begin{table}[h]
\centering
\small
\setlength{\tabcolsep}{4pt}
\resizebox{1.0\columnwidth}{!}{%
\begin{tabular}{@{}llccccccccccc|cc@{}}
\toprule
\multirow{2}{*}{\textbf{Approach}} & \multicolumn{2}{c}{\textbf{\# Train}} & \multicolumn{2}{c}{\textbf{Rest15}} & \multicolumn{2}{c}{\textbf{Rest16}} & \multicolumn{2}{c}{\textbf{FlightABSA}} & \multicolumn{2}{c}{\textbf{Coursera}} & \multicolumn{2}{c}{\textbf{Hotels}} & \multicolumn{2}{c}{\textbf{AVG}} \\
\cmidrule(lr){2-3} \cmidrule(lr){4-5} \cmidrule(lr){6-7} \cmidrule(lr){8-9} \cmidrule(lr){10-11} \cmidrule(l){12-13} \cmidrule(l){14-15}
& \textbf{TASD} & \textbf{ASQP} & \textbf{TASD} & \textbf{ASQP} & \textbf{TASD} & \textbf{ASQP} & \textbf{TASD} & \textbf{ASQP} & \textbf{TASD} & \textbf{ASQP} & \textbf{TASD} & \textbf{ASQP} & \textbf{TASD} & \textbf{ASQP} \\
\arrayrulecolor{black}\midrule
\multicolumn{15}{@{}l}{\cellcolor{gray!15}\textbf{Scenario: 0 annotated examples given}} \\
\arrayrulecolor{black}\midrule
Gemma-3-27B (0-shot) & 0 & 0 & \textbf{31.36} & \textbf{26.29} & 46.57 & \textbf{30.29} & 56.90 & 45.42 & 32.58 & 15.14 & 41.02 & 23.16 & 41.69 & 28.06 \\
\noalign{\vskip 0.1cm}              
\arrayrulecolor{lightgray}\hline  
\noalign{\vskip 0.1cm}
\rowcolor{gray!5}
LA-ABSA &  &  &  &  &  &  &  &  &  &  &  &  &  &  \\
\textit{w/ Paraphrase} & full & full & 28.54 & 20.25 & \textbf{49.68} & 28.54 & 55.69 & 46.31 & 34.06 & 15.74 & 43.22 & \textbf{27.18} & 42.24 & 27.60 \\
\rowcolor{gray!5}
\textit{w/ DLO }& full & full & 27.36 & 20.91 & 49.27 & 29.81 & \textbf{56.98} & \textbf{47.39} & \textbf{34.47} & \textbf{16.02} & \textbf{43.85} & 26.21 & \textbf{42.38} & \textbf{28.07} \\
\arrayrulecolor{black}\midrule
\multicolumn{15}{@{}l}{\cellcolor{gray!15}\textbf{Scenario: 10 annotated examples given}} \\
\arrayrulecolor{black}\midrule
Gemma-3-27B ICL & 10 & 10 & \textbf{52.66} & \textbf{40.50} & \textbf{65.19} & 47.93 & 60.87 & 45.08 & \textbf{40.37} & 21.31 & \textbf{55.17} & 28.29 & \textbf{54.85} & 36.63 \\
\noalign{\vskip 0.1cm}              
\arrayrulecolor{lightgray}\hline  
\noalign{\vskip 0.1cm}
\rowcolor{gray!5}
Paraphrase & 10 & 10 & 7.38 & 1.11 & 5.93 & 3.23 & 7.64 & 2.85 & 14.51 & 4.26 & 12.39 & 2.06 & 9.57 & 2.70 \\
DLO & 10 & 10 & 13.47 & 4.13 & 13.95 & 4.91 & 13.91 & 4.03 & 20.86 & 4.02 & 17.39 & 3.41 & 15.92 & 4.10 \\
\noalign{\vskip 0.1cm}              
\arrayrulecolor{lightgray}\hline  
\noalign{\vskip 0.1cm}
\rowcolor{gray!5}
LA-ABSA &  &  &  &  &  &  &  &  &  &  &  &  &  &  \\
\textit{w/ Paraphrase} & full & full & 46.98 & 36.43 & 61.76 & 45.91 & 58.68 & 45.93 & 38.65 & 22.99 & 53.82 & \textbf{31.07} & 51.98 & 36.47 \\
\rowcolor{gray!5}
\textit{w/ DLO }& full & full & 48.19 & 38.99 & 61.28 & \textbf{48.81} & \textbf{61.51} & \textbf{47.36} & 39.63 & \textbf{23.98} & 54.50 & 29.93 & 53.02 & \textbf{37.82} \\
\noalign{\vskip 0.1cm}              
\arrayrulecolor{lightgray}\hline  
\noalign{\vskip 0.1cm}
EDA & 110 & 110 & 28.19 & 9.53 & 19.02 & 11.69 & 19.62 & 10.37 & 26.89 & 13.07 & 25.18 & 7.27 & 23.78 & 10.39 \\
\rowcolor{gray!5}
QAIE & 26.4 & 45.2 & 16.24 & 8.19 & 10.24 & 12.84 & 14.73 & 12.42 & 22.56 & 15.81 & 17.19 & 8.19 & 16.19 & 11.49 \\
DS\textsuperscript{2}-ABSA & 21,1k & 21,1k & 28.54 & 18.29 & 30.96 & 24.38 & 26.69 & 15.73 & 18.24 & 7.81 & 31.59 & 11.54 & 27.21 & 15.55 \\
\arrayrulecolor{black}\midrule
\multicolumn{15}{@{}l}{\cellcolor{gray!15}\textbf{Scenario: 50 annotated examples given}} \\
\arrayrulecolor{black}\midrule
\rowcolor{gray!5}
Gemma-3-27B ICL & 50 & 50 & \textbf{\underline{57.16}} & 39.25 & \textbf{\underline{65.77}} & 48.06 & \textbf{62.95} & 45.25 & 39.75 & 21.71 & 56.92 & 37.17 & \textbf{56.51} & 38.29 \\
\noalign{\vskip 0.1cm}              
\arrayrulecolor{lightgray}\hline  
\noalign{\vskip 0.1cm}
Paraphrase & 50 & 50 & 34.60 & 26.62 & 34.66 & 23.25 & 31.38 & 17.42 & 31.43 & 18.21 & 36.05 & 22.59 & 33.63 & 21.62 \\
\rowcolor{gray!5}
DLO & 50 & 50 & 38.65 & 28.60 & 43.33 & 30.06 & 43.89 & 29.22 & 33.32 & 17.89 & 41.02 & 25.99 & 40.04 & 26.35 \\
\noalign{\vskip 0.1cm}              
\arrayrulecolor{lightgray}\hline  
\noalign{\vskip 0.1cm}
LA-ABSA &  &  &  &  &  &  &  &  &  &  &  &  &  &  \\
\rowcolor{gray!5}
\textit{w/ Paraphrase} & full & full & 52.50 & 37.01 & 61.12 & 46.16 & 60.76 & 46.31 & \textbf{42.50} & 24.42 & 56.40 & 40.75 & 54.65 & 38.93 \\
\textit{w/ DLO }& full & full & 55.15 & \textbf{40.53} & 60.65 & \textbf{50.06} & 62.27 & \textbf{48.95} & 42.46 & \textbf{24.50} & \textbf{57.77} & \textbf{42.11} & 55.66 & \textbf{41.23} \\
\noalign{\vskip 0.1cm}              
\arrayrulecolor{lightgray}\hline  
\noalign{\vskip 0.1cm}
\rowcolor{gray!5}
EDA & 550 & 550 & 42.49 & 30.74 & 44.66 & 34.32 & 44.54 & 35.39 & 36.02 & 23.07 & 42.45 & 31.60 & 42.03 & 31.02 \\
QAIE & 141.8 & 248.8 & 41.70 & 32.78 & 42.30 & 33.46 & 45.57 & 31.33 & 34.87 & 21.44 & 44.60 & 33.52 & 41.81 & 30.51 \\
\rowcolor{gray!5}
DS\textsuperscript{2}-ABSA & 21,7k & 21,6k & 38.08 & 30.01 & 43.94 & 36.67 & 34.63 & 24.17 & 29.71 & 16.81 & 38.54 & 23.23 & 36.98 & 26.18 \\
\arrayrulecolor{black}\midrule
\multicolumn{15}{@{}l}{\cellcolor{gray!15}\textbf{Full set of human-annotated examples: SOTA approaches}} \\
\arrayrulecolor{black}\midrule
Paraphrase & full & full & - & 47.72 & - & 59.30 & 69.26 & 58.17 & 51.02 & 32.63 & 67.01 & 55.19 & 62.43 & 50.60 \\
\rowcolor{gray!5}
DLO & full & full & - & \textbf{\underline{49.33}} & - & \textbf{\underline{61.80}} & \textbf{\underline{69.30}} & \textbf{\underline{60.10}} & \textbf{\underline{52.38}} & \textbf{\underline{33.07}} & \textbf{\underline{68.71}} & \textbf{\underline{56.56}} & \textbf{\underline{63.46}} & \textbf{\underline{52.17}} \\
\bottomrule
\end{tabular}
}
\caption{\textbf{Recall scores of LA-ABSA}. Results are evaluated against EDA-based data augmentation methods, QAIE, DS\textsuperscript{2}-ABSA, and prompting baselines (0, 10, and 50 annotated examples) as reported by \citet{hellwig2025we}, as well as fully supervised models including DLO \citep{hu2022improving} and Paraphrase \citep{zhang2021aspect}. The highest recall scores within each annotation regime (0, 10, 50 or all examples) are shown in bold; the best overall scores across all settings are underlined.}

\label{tab:results-recall}
\end{table}

\clearpage

\subsection{F1 Macro}
\begin{table}[h]
\centering
\small
\setlength{\tabcolsep}{4pt}
\resizebox{1.0\columnwidth}{!}{%
\begin{tabular}{@{}llccccccccccc|cc@{}}
\toprule
\multirow{2}{*}{\textbf{Approach}} & \multicolumn{2}{c}{\textbf{\# Train}} & \multicolumn{2}{c}{\textbf{Rest15}} & \multicolumn{2}{c}{\textbf{Rest16}} & \multicolumn{2}{c}{\textbf{FlightABSA}} & \multicolumn{2}{c}{\textbf{Coursera}} & \multicolumn{2}{c}{\textbf{Hotels}} & \multicolumn{2}{c}{\textbf{AVG}} \\
\cmidrule(lr){2-3} \cmidrule(lr){4-5} \cmidrule(lr){6-7} \cmidrule(lr){8-9} \cmidrule(lr){10-11} \cmidrule(l){12-13} \cmidrule(l){14-15}
& \textbf{TASD} & \textbf{ASQP} & \textbf{TASD} & \textbf{ASQP} & \textbf{TASD} & \textbf{ASQP} & \textbf{TASD} & \textbf{ASQP} & \textbf{TASD} & \textbf{ASQP} & \textbf{TASD} & \textbf{ASQP} & \textbf{TASD} & \textbf{ASQP} \\
\arrayrulecolor{black}\midrule
\multicolumn{15}{@{}l}{\cellcolor{gray!15}\textbf{Scenario: 0 annotated examples given}} \\
\arrayrulecolor{black}\midrule
Gemma-3-27B (0-shot) & 0 & 0 & \textbf{33.68} & \textbf{23.64} & \textbf{45.94} & 23.64 & \textbf{55.24} & \textbf{39.50} & \textbf{33.78} & \textbf{13.23} & \textbf{29.70} & \textbf{18.49} & \textbf{39.67} & \textbf{23.70} \\
\noalign{\vskip 0.1cm}              
\arrayrulecolor{lightgray}\hline  
\noalign{\vskip 0.1cm}
\rowcolor{gray!5}
LA-ABSA &  &  &  &  &  &  &  &  &  &  &  &  &  &  \\
\textit{w/ Paraphrase} & full & full & 22.88 & 19.93 & 35.84 & 26.20 & 45.95 & 33.61 & 11.85 & 4.09 & 16.20 & 12.44 & 26.54 & 19.25 \\
\rowcolor{gray!5}
\textit{w/ DLO }& full & full & 24.12 & 21.02 & 36.12 & \textbf{29.16} & 45.37 & 36.32 & 12.10 & 3.70 & 17.18 & 13.10 & 26.98 & 20.66 \\
\arrayrulecolor{black}\midrule
\multicolumn{15}{@{}l}{\cellcolor{gray!15}\textbf{Scenario: 10 annotated examples given}} \\
\arrayrulecolor{black}\midrule
Gemma-3-27B ICL & 10 & 10 & \textbf{54.53} & \textbf{\underline{38.67}} & \textbf{64.33} & \textbf{43.21} & \textbf{56.32} & 36.58 & \textbf{\underline{41.75}} & \textbf{\underline{20.44}} & \textbf{31.15} & 18.05 & \textbf{49.62} & \textbf{31.39} \\
\noalign{\vskip 0.1cm}              
\arrayrulecolor{lightgray}\hline  
\noalign{\vskip 0.1cm}
\rowcolor{gray!5}
Paraphrase & 10 & 10 & 3.47 & 0.46 & 2.59 & 0.87 & 2.38 & 1.03 & 2.17 & 0.66 & 2.65 & 0.47 & 2.65 & 0.70 \\
DLO & 10 & 10 & 6.10 & 1.42 & 6.01 & 1.25 & 4.74 & 1.35 & 3.33 & 0.60 & 4.10 & 0.83 & 4.86 & 1.09 \\
\noalign{\vskip 0.1cm}              
\arrayrulecolor{lightgray}\hline  
\noalign{\vskip 0.1cm}
\rowcolor{gray!5}
LA-ABSA &  &  &  &  &  &  &  &  &  &  &  &  &  &  \\
\textit{w/ Paraphrase} & full & full & 34.17 & 23.98 & 48.59 & 34.59 & 51.20 & 36.29 & 17.01 & 8.54 & 26.90 & \textbf{19.19} & 35.57 & 24.52 \\
\rowcolor{gray!5}
\textit{w/ DLO }& full & full & 35.03 & 26.77 & 50.39 & 36.25 & 53.19 & \textbf{39.01} & 17.14 & 9.38 & 29.02 & 18.18 & 36.95 & 25.92 \\
\noalign{\vskip 0.1cm}              
\arrayrulecolor{lightgray}\hline  
\noalign{\vskip 0.1cm}
EDA & 110 & 110 & 11.33 & 5.57 & 7.94 & 2.22 & 8.05 & 3.64 & 4.22 & 2.02 & 5.49 & 2.27 & 7.40 & 3.15 \\
\rowcolor{gray!5}
QAIE & 26.4 & 45.2 & 14.86 & 5.37 & 7.84 & 14.29 & 12.02 & 7.00 & 4.30 & 1.90 & 3.52 & 2.86 & 8.51 & 6.28 \\
DS\textsuperscript{2}-ABSA & 21,1k & 21,1k & 18.83 & 12.17 & 26.68 & 15.60 & 22.66 & 13.04 & 9.65 & 3.17 & 16.97 & 7.89 & 18.96 & 10.37 \\
\arrayrulecolor{black}\midrule
\multicolumn{15}{@{}l}{\cellcolor{gray!15}\textbf{Scenario: 50 annotated examples given}} \\
\arrayrulecolor{black}\midrule
\rowcolor{gray!5}
Gemma-3-27B ICL & 50 & 50 & \textbf{\underline{56.01}} & \textbf{36.01} & \textbf{\underline{64.81}} & \textbf{\underline{44.56}} & \textbf{54.06} & 36.67 & \textbf{38.24} & \textbf{17.80} & \textbf{35.47} & \textbf{25.24} & \textbf{\underline{49.72}} & \textbf{32.06} \\
\noalign{\vskip 0.1cm}              
\arrayrulecolor{lightgray}\hline  
\noalign{\vskip 0.1cm}
Paraphrase & 50 & 50 & 19.75 & 14.05 & 16.35 & 8.35 & 12.43 & 7.23 & 9.59 & 2.99 & 13.87 & 7.54 & 14.40 & 8.03 \\
\rowcolor{gray!5}
DLO & 50 & 50 & 27.78 & 18.10 & 22.46 & 13.05 & 28.90 & 19.71 & 8.68 & 3.11 & 16.50 & 9.74 & 20.86 & 12.74 \\
\noalign{\vskip 0.1cm}              
\arrayrulecolor{lightgray}\hline  
\noalign{\vskip 0.1cm}
LA-ABSA &  &  &  &  &  &  &  &  &  &  &  &  &  &  \\
\rowcolor{gray!5}
\textit{w/ Paraphrase} & full & full & 35.74 & 26.57 & 48.34 & 31.01 & 52.88 & 36.65 & 17.96 & 7.37 & 28.20 & 19.79 & 36.62 & 24.28 \\
\textit{w/ DLO }& full & full & 41.64 & 28.42 & 49.37 & 37.37 & 52.17 & \textbf{40.88} & 18.07 & 8.10 & 30.19 & 22.59 & 38.29 & 27.47 \\
\noalign{\vskip 0.1cm}              
\arrayrulecolor{lightgray}\hline  
\noalign{\vskip 0.1cm}
\rowcolor{gray!5}
EDA & 550 & 550 & 27.35 & 17.61 & 22.38 & 16.12 & 33.71 & 24.74 & 12.22 & 5.72 & 15.85 & 13.07 & 22.30 & 15.45 \\
QAIE & 141.8 & 248.8 & 44.81 & 30.84 & 45.50 & 32.23 & 37.41 & 22.29 & 20.79 & 10.20 & 24.19 & 18.51 & 34.54 & 22.82 \\
\rowcolor{gray!5}
DS\textsuperscript{2}-ABSA & 21,7k & 21,6k & 24.95 & 18.14 & 31.43 & 21.81 & 28.52 & 17.92 & 11.86 & 4.69 & 18.35 & 13.06 & 23.02 & 15.12 \\
\arrayrulecolor{black}\midrule
\multicolumn{15}{@{}l}{\cellcolor{gray!15}\textbf{Full set of human-annotated examples: SOTA approaches}} \\
\arrayrulecolor{black}\midrule
Paraphrase & full & full & - & - & - & - & \textbf{\underline{58.48}} & 48.47 & 30.67 & \textbf{15.66} & 37.49 & 30.31 & 42.21 & 31.48 \\
\rowcolor{gray!5}
DLO & full & full & - & - & - & - & 57.24 & \textbf{\underline{49.41}} & \textbf{32.25} & 14.91 & \textbf{\underline{38.14}} & \textbf{\underline{33.04}} & \textbf{42.54} & \textbf{\underline{32.45}} \\
\bottomrule
\end{tabular}
}
\caption{\textbf{Macro-averaged F1 scores of LA-ABSA.} Results are evaluated against EDA-based data augmentation methods, QAIE, DS\textsuperscript{2}-ABSA, and prompting baselines (0, 10, and 50 annotated examples) as reported by \citet{hellwig2025we}, as well as fully supervised models including DLO \citep{hu2022improving} and Paraphrase \citep{zhang2021aspect}. The highest F1 macro scores within each annotation regime (0, 10, 50 or all examples) are shown in bold; the best overall scores across all settings are underlined.}

\label{tab:results-f1-macro}
\end{table}